\documentclass[letterpaper, 10 pt, conference]{ieeeconf}  % Comment this line out if you need a4paper

\IEEEoverridecommandlockouts                              % This command is only needed if 
\usepackage{amsmath}
\usepackage{nicefrac}
\usepackage{units}
\usepackage[pdftex]{graphicx}
\usepackage{import}
\usepackage{xcolor}
\usepackage{colortbl}
\usepackage{caption}
\title{\LARGE \bf
Considering Human Factors in Risk Maps \\ for Robust and Foresighted Driver Warning
}

\author{Tim Puphal$^1$, Ryohei Hirano$^2$, Malte Probst$^1$, Raphael Wenzel$^1$ and Akihito Kimata$^2$% <-this % stops a space
\thanks{$^1$ Honda Research Institute Europe, Carl-Legien-Str. 30, 63073 Offenbach, Germany. Email: {\tt\footnotesize \{firstname.lastname\}@honda-ri.de}}
\thanks{
$^2$ Safety and Human Factors Research, Honda R$\&$D Co., Ltd. 4630 Shimotakanezawa, 321-3393 Tochigi, Japan. Email: {\tt\footnotesize \{firstname$\_$lastname\}@jp.honda}
}}

\begin{document}

\maketitle
\thispagestyle{empty}
\pagestyle{empty}

%%%%%%%%%%%%%%%%%%%%%%%%%%%%%%%%%%%%%%%%%%%%%%%%%%%%%%%%%%%%%%%%%%%%%%%%%%%%%%%%
\begin{abstract}
Driver support systems that include human states in the support process is an active research field. Many recent approaches allow, for example, to sense the driver's drowsiness or awareness of the driving situation. However, so far, this rich information has not been utilized much for improving the effectiveness of support systems. In this paper, we therefore propose a warning system that uses human states in the form of driver errors and can warn users in some cases of upcoming risks several seconds earlier than the state of the art systems not considering human factors. The system consists of a behavior planner Risk Maps which directly changes its prediction of the surrounding driving situation based on the sensed driver errors. By checking if this driver's behavior plan is objectively safe, a more robust and foresighted driver warning is achieved. In different simulations of a dynamic lane change and intersection scenarios, we show how the driver's behavior plan can become unsafe, given the estimate of driver errors, and experimentally validate the advantages of considering human factors. 
\end{abstract}

\section{Introduction}
Support systems are successfully helping users in many everyday driving situations. State-of-the-art support systems are, for example, adaptive cruise control, collision mitigation systems or parking assistance \cite{bengler2014}. Such systems successfully help to ensure safety and comfort for the driver. For more improved driver support shaped specifically for the user, the human state is recently included increasingly in the process. Adaptive cruise control, for example, checks if the driver has its hands on the steering wheel to constrain the system use \cite{bianchi2013}. Recently, in human factors research, technologies are additionally developed that estimate the user's gaze, posture, drowsiness, and general awareness of the driving situation \cite{zhu2021}. However, not many works have so far sufficiently utilized the human state information for improving the effectiveness of the support system. Leveraging human factors in driver support, particularly in the therein contained risk models and behavior planners, constitutes a large opportunity. 

An example of a driving situation, where human factors improve user support is given in Fig. 1. The figure shows a motorcycle which intends to change lane because of a slower car in the front. Since there is a further car in the neighboring lane, the motorcycle needs to specifically consider this agent in the behavior planning. We further assume that the rider of the motorcycle has a human state that can constitute three different driver errors: a) notice error, b) forecast error and c) inference error. For the notice error, the rider will not be aware of the car in the neighboring lane and might make a sudden lane change. In the same line, in the case of a forecast error, the motorcycle rider estimates the velocity of the neighboring car wrong and could misunderstand the time gap for the lane change and finally, in the case of an inference error, the rider predicts a wrong intention for the neighboring car, e.g., that the car will also change lane, which would also lead to a different view compared to what is happening in the real world. These errors highly influence the driving situation outcome.

\begin{figure}[t!]
  \centering
  \vspace*{0.06cm}
  \resizebox{0.96\linewidth}{!}{%% Creator: Inkscape inkscape 0.92.4, www.inkscape.org
%% Accompanies image file '1_intro_human_factors2.pdf' (pdf, eps, ps)
\begingroup%
  \makeatletter%
  \providecommand\color[2][]{%
    \errmessage{(Inkscape) Color is used for the text in Inkscape, but the package 'color.sty' is not loaded}%
    \renewcommand\color[2][]{}%
  }%
  \providecommand\transparent[1]{%
    \errmessage{(Inkscape) Transparency is used (non-zero) for the text in Inkscape, but the package 'transparent.sty' is not loaded}%
    \renewcommand\transparent[1]{}%
  }%
  \providecommand\rotatebox[2]{#2}%
  \newcommand*\fsize{\dimexpr\f@size pt\relax}%
  \newcommand*\lineheight[1]{\fontsize{\fsize}{#1\fsize}\selectfont}%
  \ifx\svgwidth\undefined%
    \setlength{\unitlength}{300.21773499bp}%
    \ifx\svgscale\undefined%
      \relax%
    \else%
      \setlength{\unitlength}{\unitlength * \real{\svgscale}}%
    \fi%
  \else%
    \setlength{\unitlength}{\svgwidth}%
  \fi%
  \global\let\svgwidth\undefined%
  \global\let\svgscale\undefined%
  \makeatother%
  \begin{picture}(1,0.40648339)%
    \lineheight{1}%
    \setlength\tabcolsep{0pt}%
    \put(0.67418551,0.1517885){\color[rgb]{0,0,0}\makebox(0,0)[lt]{\lineheight{1.25}\smash{\begin{tabular}[t]{l}notice error\end{tabular}}}}%
    \put(0.67418551,0.10632805){\color[rgb]{0,0,0}\makebox(0,0)[lt]{\lineheight{1.25}\smash{\begin{tabular}[t]{l}forecast error\end{tabular}}}}%
    \put(0.67418551,0.05882703){\color[rgb]{0,0,0}\makebox(0,0)[lt]{\lineheight{1.25}\smash{\begin{tabular}[t]{l}inference error\end{tabular}}}}%
    \put(0.060828958,0.38512901){\color[rgb]{0,0,0}\makebox(0,0)[lt]{\lineheight{1.25}\smash{\begin{tabular}[t]{l}real world\end{tabular}}}}%
    \put(0.079128958,0.10763029){\color[rgb]{0,0,0}\makebox(0,0)[lt]{\lineheight{1.25}\smash{\begin{tabular}[t]{l}driver's\\view\end{tabular}}}}%
    \put(0,0){\includegraphics[width=\unitlength,page=1]{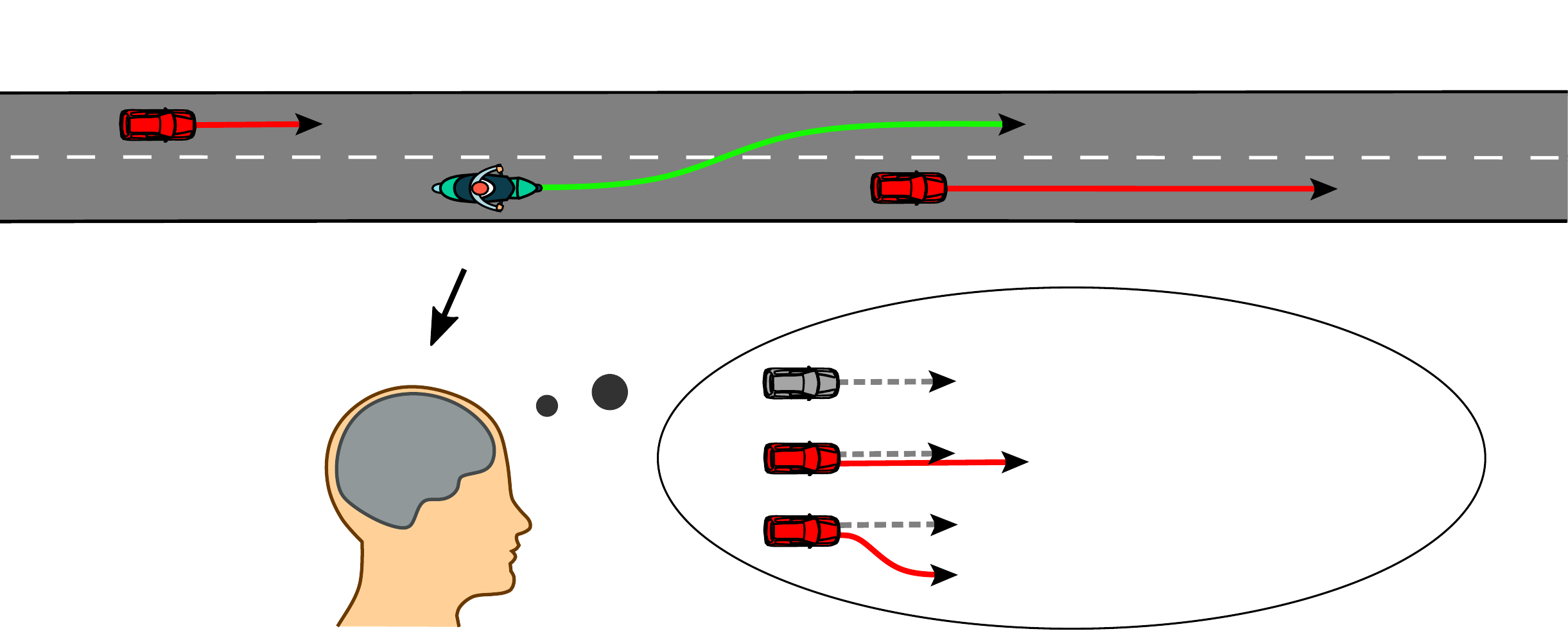}}%
  \end{picture}%
\endgroup%
}
  \vspace*{0.05cm}
  \caption[]{The image shows an example driving situation of a dynamic lane change and how different driver errors influence the motorcycle rider's view of the situation. In this paper, we propose a warning system that can warn users more robust and earlier considering these driver errors.}
  \label{fig:filtering_toy_example}
\end{figure}
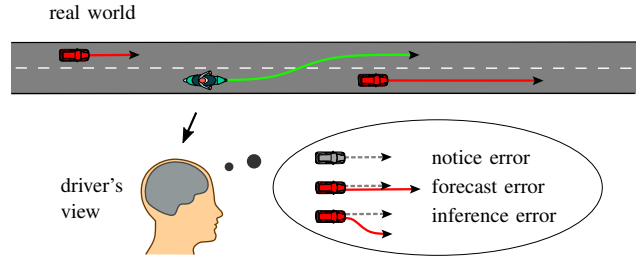

Building upon the definition of driver errors\footnote{Please note that for simplicity, we will sometimes use the word driver for the driver of a car, the rider of a motorcycle or operator of any vehicle.}, in this paper, we propose a warning system concept that in some cases is able to warn users of upcoming risks several seconds earlier than state of the art not considering human factors. The system consists of a behavior planner which changes the prediction of the surrounding driving situation based on human factors, here in the form of the sensed driver errors from the human state. This enables to plan a motion how the driver perceives the environment. By comparing this driver's view with the real environment, a more robust and foresighted driver warning can be achieved. For the behavior planner, we use a model-based planner, called Risk Maps, because of its explainability of the future driving risks. In simulation experiments of variations for the dynamic lane change from Fig. 1 and of further intersection scenarios, we will show how the driver's motion plan can become unsafe, given an estimate of the driver errors, and show the effectiveness of a more proactive driver warning. 

\subsection{Related Work}

In previous work, so-called personalization of driver support systems in order to bring the system closer to users has been given much attention. A survey of different approaches for personalization is, for example, given in the work of \cite{hasenjaeger2017}. As a concrete example, the authors of \cite{siebinga2022} adapt parameters of a cost function for a behavior planner in order to better match the system's output to the driver`s behavior. In this context, they use inverse reinforcement learning. While personalization allows for less false positive warnings, the time of the warning is, however, not much improved.

Another larger body of work is the investigation of perceived risks. While personalization of support systems is an interesting topic for computer scientists, the human factors research community has worked on trying to quantify the risk of driving situations from the human perspective. This is similar to used risk models for behavior planners of self-driving cars, such as from \cite{puphal2019}, however, different human-focused risk sources are investigated that might not be considered usually. For example, the authors of \cite{kolekar2021} investigated the general driver`s feeling of distance to the curb sides, the authors of \cite{charlton2014} examined the perceived risk from the visibility of driving situations from images and the authors of \cite{fridman2018} detect if the driver is not looking on the road but on the smartphone which is a further factor leading to driving risks. Approaches for perceived risk help to understand the lack of human factors for support systems, but to the knowledge of the authors, these approaches did not incorporate how driver errors, such as the wrong prediction of surrounding objects, influence the driving risk and behavior plan in order to improve support systems.

Finally, there are few works which take the human driver dynamically in the loop for support systems, similar to the presented work in this paper. For example, the work of \cite{wang2020} allows users to help the system better predict lane changes of other vehicles. The user can see the current system prediction and over speech input, can correct the prediction. After many attempts, the system learns to better warn the driver, however this work takes the opposite direction, which needs the driver to first improve the system than the system being intelligent and warning the driver earlier from the start. Another interesting work is described in \cite{mccall2007}. The authors check if the driver intends to brake with a camera and if the driver does brake, the system will not warn of upcoming critical objects. This improved brake assist dynamically improves the driver support but is limited in scope.

\subsection{Contribution} 
In this paper, we build upon related work and incorporate driver errors into a warning system to inform the driver of its wrong perception. This paradigm is different to current support systems in production which only take into account the real situation and different to previous attempts of considering human factors through personalization or further perceived risk sources. To the knowledge of the authors, this is the first time such a concept is proposed. The contributions of this paper are two-fold: %lie in the intersection of human factors models and motion or behavior planning and are
\begin{enumerate}
 \item We propose a general interface of human factors to risk models and behavior planners with driver errors. 
 \item We develop a warning system concept that compares the driver's view including the driver errors with the real world not including the driver errors to warn users more robust and foresighted.
\end{enumerate}

The remainder of the paper is structured accordingly. In the next Section \ref{sec:human_factors}, the driver errors and how they influence the behavior planner Risk Maps are explained in more detail. Section \ref{sec:warning_system} then describes the structure of the novel warning system concept using two Risk Maps: the perceived Risk Maps and objective Risk Maps. Finally, Section \ref{sec:experiments} will give examples for each driver error and the performance of the warning system and Section \ref{sec:conclusion} outlines the conclusion and outlook.

\section{Human Factors for Risk Maps}
\label{sec:human_factors}

The focus of this paper lies in the definition, understanding and usage of driver errors for the perception of risk. In this section, we will therefore define the interface of driver errors for Risk Maps and show qualitatively the influence of driver errors on the perceived driver view of the situation. Herein, we will assume a state estimation module that senses human states and accordingly driver errors from vehicle sensors as given. For approaches of state estimation modules, please refer, for example, to the work of \cite{fridman2018}. Estimating the driver errors from sensors is another research work by itself.

\subsection{Driver Errors}
In the following, we propose interfaces for the three driver errors: a) notice error, b) forecast error and c) inference error. A notice error depicts the condition that the driver is unaware of another object. If the driver has not seen the other object, or saw it more than a fixed time ago, the driver might not be aware of this other object. %The discovery error estimation might also be enhanced based on the drowsiness of the driver. 
We define the interface for notice error as 
\begin{equation}
o_{\text{per}}=
    \begin{cases}
        \text{aware,} \hspace{1.074cm} \text{for } \text{NE} \in [0, 0.5), \\
        \text{not aware,} \hspace{0.5cm} \text{for } \text{NE} \in [0.5, 1], \\
    \end{cases}
\label{eq:DE}
\end{equation}
with the discrete variable $o_{\text{per}}$ describing the perceived awareness for the object and an estimated value of the notice error \text{NE} from a state estimation module. The driver can be either aware or not aware of the other object. The variables and all following variables are here dynamic and given for the current time $t$, which gives us $o_{\text{per}}(t)$ and $\text{NE}(t)$. % All following errors are dependent on the current time as well.

A forecast error represents the condition that the driver estimates the velocity of another vehicle incorrectly. We thus define the interface of the forecast error as 
\begin{equation}
v_\text{per} = v_{\text{obj}} + \text{FE} \cdot v_{\text{off}}, \hspace{0.5cm}\text{for FE} \in [-1, 1],
\end{equation}
in which the therein contained $v_{\text{per}}$ is the perceived velocity of another vehicle for the ego driver and $\text{FE}$ represents the forecast error from a state estimation module. This error defines the factor of how much an offset value $v_{\text{off}}$ changes the correct, objective velocity $v_{\text{obj}}$, whereby $v_{\text{off}}$ is a parameter and can be chosen depending on the user.  

Finally, an inference error describes a driver that predicts the future driving path of another vehicle wrong. We define the interface of the inference error as 
\begin{equation}
\mathbf{p}_\text{per} =     
\begin{cases}
    \mathbf{p}_{\text{obj}}, \hspace{0.616cm} \text{IE} \in [0, 0.5), \\
    \mathbf{p}_{\text{pred}}, \hspace{0.5cm} \text{IE} \in [0.5, 1]. \\
\end{cases}
\end{equation}
In the equation, two paths are here given. First, the correct, objective future path of the other vehicle $\mathbf{p}_{\text{obj}}$ and second, the predicted path from the ego driver for the other vehicle $\mathbf{p}_{\text{pred}}$. Depending on the estimated value of the inference error $\text{IE}$ from a state estimation module, the perceived path from the driver $\mathbf{p}_{\text{per}}$ might be wrong. 

One important aspect in the aformentioned definitions of the driver errors is that, in the end, the state estimation should output an error from the perspective of the ego driver for each other vehicle $j$. All three errors are thus defined as $\text{NE}_j$, $\text{FE}_j$, $\text{IE}_j$ with $j$ being the other considered vehicle. 

\subsection{Influence on Risk Maps}
\label{sec:perceived_riskmaps}

We target to use the aformentioned interfaces and input driver errors into Risk Maps to achieve a visualization of the perceived risk.
Risk Maps is thereby a driver behavior planner that models risk for different possible behaviors of the driver over the future time. The planner uses a stochastic risk model that can estimate, amongst others, a collision risk between two agents as the product of its probability and severity. 
As one representation, the planned velocity over the future time can be taken. Risk Maps is then using a function of risk over the velocity-time space and plans behaviors with low risk, such as braking to avoid an object. 

\begin{figure}[t!]
  \centering
  \vspace*{0.7cm}
  \hspace{0.1cm}
  \resizebox{0.94\linewidth}{!}{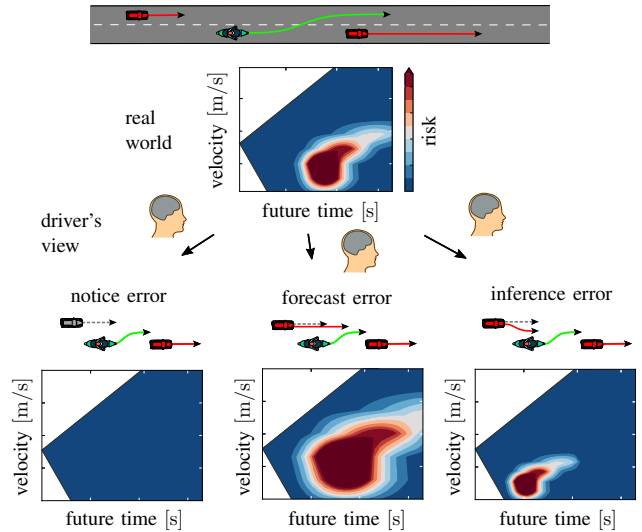}
  \vspace*{-0.05cm}
  \caption[]{Visualization of the perceived risk using Risk Maps for the three driver errors of notice error, forecast error and inference error. Depending on the error values, the driver's risk perception can change.} 
  \vspace*{-0.1cm}
  \label{fig:perceived_riskmaps}
\end{figure}

Fig.~\ref{fig:perceived_riskmaps} shows an example of using Risk Maps for the lane change scenario from the introduction. On the top, the figure shows a plot of possible collision risks with the back car from the perspective of the motorcycle rider when making a lane change. If the rider accelerates, the rider reaches the upper plot area (large velocity) and would encounter no risk. However, if the rider plans to brake, the rider will reach the lower plot area (small velocity) and cause risk because the motorcycle will collide with the back car. This plot visualizes the real world based on the sensed vehicle velocities. 

We add the driver errors in Risk Maps by also inputting the interface variables $o_{\text{per}}, v_{\text{per}}$, and $\mathbf{p}_{\text{per}}$ from the last section. This is the main difference to the original Risk Maps behavior planner which was presented in \cite{puphal2022}. On the bottom, Fig. \ref{fig:perceived_riskmaps} depicts how the visualization of Risk Maps changes for the three driver errors of notice error, forecast error and inference error. 

In the scenario, the motorcycle rider makes a lane change and should predict both agents to keep driving on their lane. If the rider holds driver errors, the rider might perceive this situation wrong. In the case of a notice error, the rider will not be aware of the other car in the back. In this example, there is no risk in the plot of Risk Maps. The perceived risk is therefore lower. If there is a forecast error, the motorcycle rider will, for example, estimate the back car's velocity too high and consequently, the perceived risk is higher. Lastly, if there is an inference error, the rider will assume the back car to change lanes. In this case, if the rider also makes a lane change, both will avoid each other and the perceived risk is lower. In the example descriptions, the rider has here only driver errors for the back car.

\section{Warning System}
\label{sec:warning_system}

In the last section, we described the influence of human factors in the form of driver errors for the perceived risk of the driver  using Risk Maps. However, we target to use the human factors information to warn and support the driver. In this section, we therefore describe how we estimate a driver behavior based on the perceived Risk Maps and evaluate driver errors.

\vspace{-0.05cm}
\subsection{Evaluating Driver Errors}

\begin{figure}[t!]
  \centering
  
  \vspace{-0.1cm}
  \resizebox{0.92\linewidth}{!}{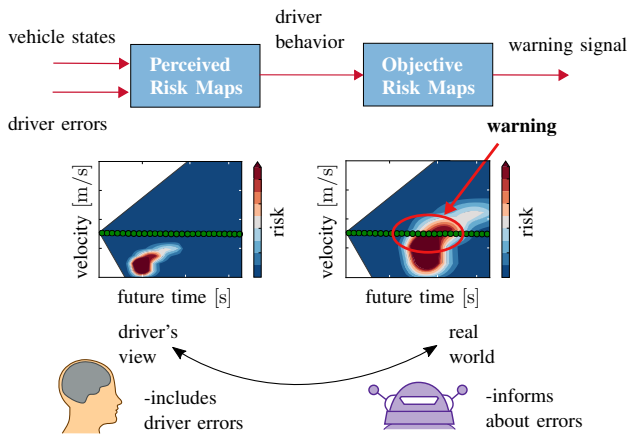}
  
  \vspace*{0.17cm}
  
  \caption[]{\hspace{0.07cm} In the novel warning system concept, the perceived Risk Maps with driver errors is compared with the objective Risk Map without driver errors in order to derive a warning signal and warn the driver better. This allows to include human factors in the driver support.} 
  \vspace{-0.03cm}
  \label{fig:warning_system}
\end{figure}

A block diagram of the novel warning system is illustrated in Fig. \ref{fig:warning_system}. In an initial step, the driver behavior is estimated with the perceived Risk Maps module that inputs the driver errors. Risk Maps allows to plan a safe behavior by finding a velocity profile that avoids the risk spots in the visualization. This can be achieved, for example, by sampling different target velocities that constitute acceleration and deceleration driver behaviors and taking the target velocity that minimizes total costs $C$ consisting of risk $R$, utility $U$ and comfort $O$. This can be written as follows
\begin{align}
C(v^h) = R(v^h) - U(v^h) &+ O(v^h), \\ \text{with } 
v_{\text{tar}} = \text{argmin}_{h} C(v^h)
\end{align}
The result is a found driver behavior $v_{\text{tar}}$ with minimal costs 

\noindent among different sampled velocities $v^h$. In Fig. \ref{fig:warning_system}, the found velocity profile is visualized with a green curve and, in the example, the target velocity is to drive with constant velocity in order to avoid the risk spot in the lower area. %in order to avoid the risk spot.

Since we use perceived Risk Maps in order to find a safe behavior, this driver behavior is resulting from the driver's view of the driving situation and is actually only safe for the driver based on the current human factors input. To retrieve a warning output for the driver, we further take this seemingly safe ego behavior and evaluate its resulting risk without human factors input. This module is called objective Risk Maps and we define the warning signal as
\begin{equation}
 W(t) = R_{\text{obj}}(t)
\end{equation}
with $R_{\text{obj}}$ as the objective risk value of the driver behavior and with $t$ as the current time.\footnote{Remark: It is not necessary to compute the whole plot of the objective Risk Maps in order to warn the driver. One single evaluation of the risk for the found driver behavior $R_{\text{obj}}$ is sufficient to warn the user. However, for explainability, we compute the whole plot in this paper.}

The risk model used for the perceived Risk Maps and the objective Risk Maps includes a Gaussian model that models 2D Gaussian distributions around the positions of the vehicles and estimates the collision risk between vehicles by taking the overlap of the Gaussians. For a description of the complete risk model, please refer to the work \cite{puphal2022}. 

%\subsection{Evaluating Driver Errors}

Fig. \ref{fig:warning_system} explains the overall warning concept visually. Perceived Risk Maps embodies the human driver that includes the driver errors and objective Risk Maps embodies the machine supporting the user that does not include the driver errors and informs about errors. The green curve depicting the planned driver behavior is safe for the perceived Risk Maps but goes through the risk spot in the objective Risk Maps plot. The driver is perceiving the driving situation wrong and a warning is issued. While the perceived risk of the found driver behavior $R_{\text{per}}$ is low, the objective risk $R_{\text{obj}}$ is high and critical. 

In this sense, only when the values of $R_{\text{per}}(t)$ and $R_{\text{obj}}(t)$ are different, the warning signal becomes critical. In other words, if the driver's situational view differs from the real world, the system warns the driver. We will show in the next section that this dependency reduces false positive warnings for the warning system. Moreover, by inputting additional information into the system with driver errors, the system is able to find risks that are otherwise not detected or detected later. Using human factors in the form of driver errors has the potential to warn robust and foresighted.

\vspace{-0.07cm}
\section{Experiments}
\label{sec:experiments}

The following section describes the experiments that show improved driver support using the presented warning system concept of this paper. The experiments were done with an own driving simulator written in python that allows to input driver errors for one agent, which is considered as the ego vehicle and which is supported by the warning system. All other agents are simulated with fixed behaviors. Here, the simulator uses kinematics models for  
updating the positions of the agents.

\begin{figure*}[t!]
  \centering
  
  \vspace{0.15cm}
  \resizebox{0.98\linewidth}{!}{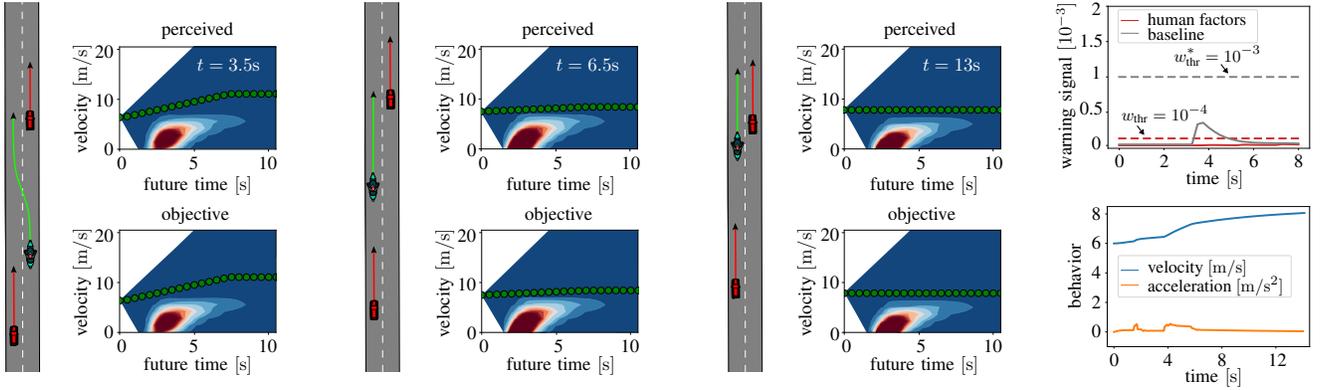}
  \vspace{0.02cm}    
  \caption[]{An example of the dynamic lane change scenario with the rider of the motorcycle having no driver error. The rider is making a lane change successfully and slightly accelerates. Since both perceived and objective Risk Maps are the same, the warning signal stays low.} 
  \vspace{0.05cm}
  \label{fig:warning_no_driver_error}
\end{figure*}
\begin{figure*}[t!]
  \centering
  \resizebox{0.98\linewidth}{!}{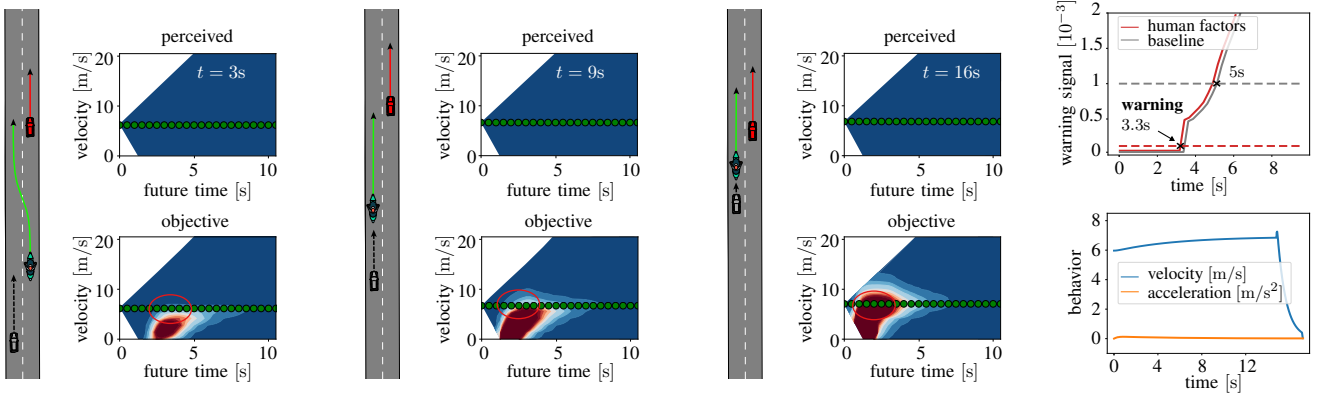}
  \vspace{0.02cm}  
  \caption[]{Example with rider having notice error. The rider is not aware of the back car and a collision happens at the end of the simulation. Since the found driving behavior from the perceived Risk Maps goes through the risk spot of the objective Risk Maps plot, a warning is issued. Using human factors allows to warn the rider around 2 seconds earlier than the baseline (see warning signal plot).} 
  \vspace{0.05cm}
  \label{fig:warning_discovery_error}
\end{figure*}
\begin{figure*}[t!]
  \centering
  \resizebox{0.98\linewidth}{!}{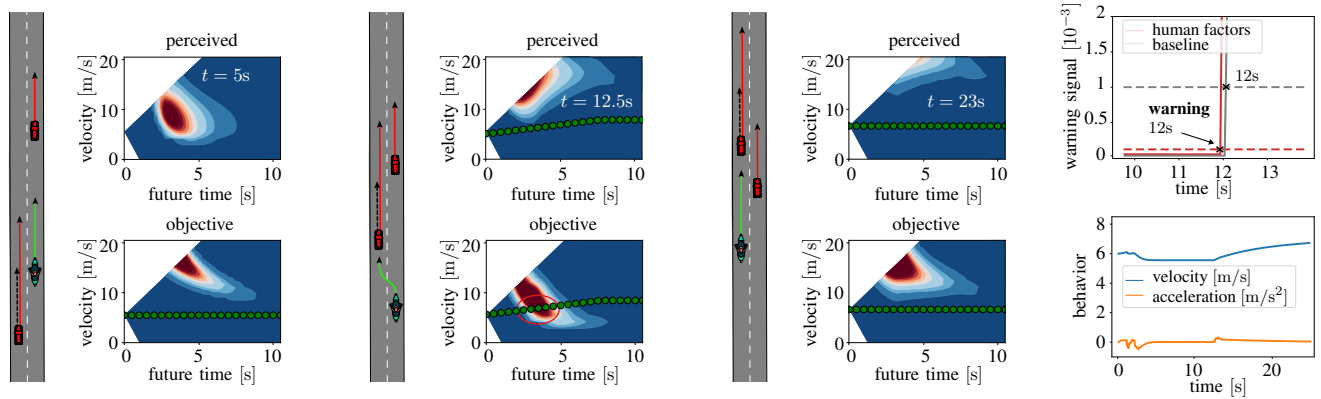}
  \caption[]{Example with rider having forecast error. The rider overestimates the speed of the back car and does not dare to make a lane change because the risk spot is wide (see perceived Risk Maps plot of first timestep). The rider is warned about the wrong perception in the course of the simulation.} 
  \vspace{0.1cm}
  \label{fig:warning_estimation_error}
\end{figure*}

\begin{figure*}[t!]
  \centering
  \vspace{0.28cm}
  \resizebox{0.98\linewidth}{!}{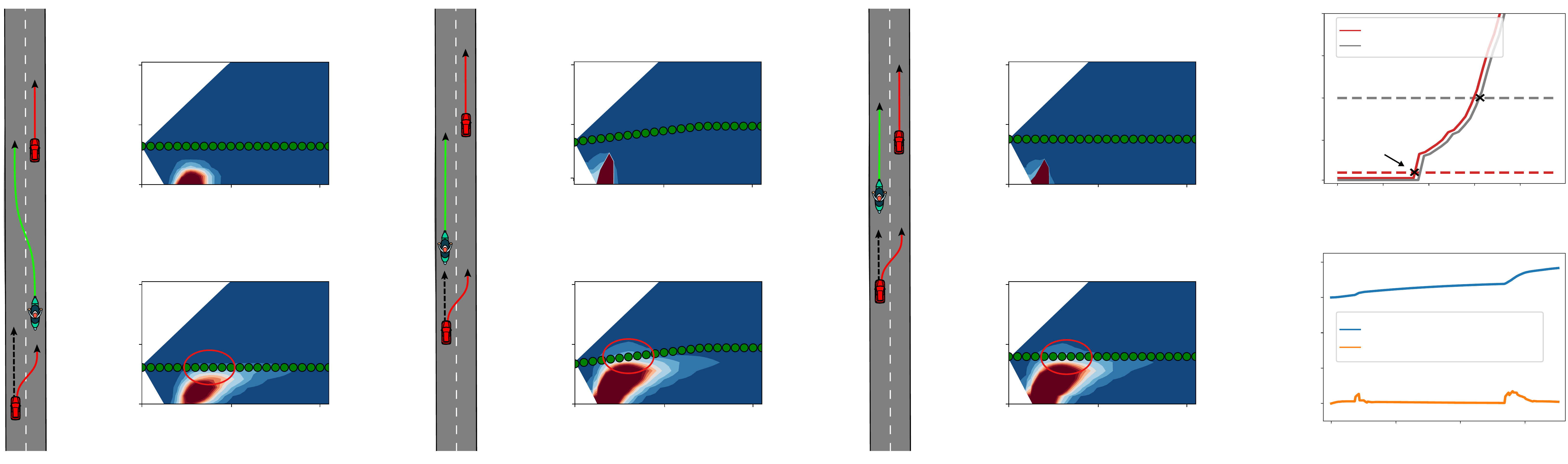}
  
  \caption[]{Example with rider having inference error. The rider wrongly predicts a lane change for the back car. The found driving behavior is dangerous (see objective Risk Maps visualizations) and the warning signal is high as for the other examples with driver errors. The rider is warned around 3 seconds earlier than state of the art.}
  \vspace{-0.1cm}
  \label{fig:warning_prediction_error}
\end{figure*}

%intersection scenarios
This section will first show simulations of variations from all driver errors (i.e., no driver error, notice error, forecast error and inference error) for the dynamic lane change scenario that was outlined in the introduction. Afterwards, tests of the warning system will be given for two different intersection scenarios that also include driver errors. Note that in the simulations, the driver error inputs are assumed given (i.e., no state estimation module is used). 

In order to show the warning improvement using human factors, we compare in the tests the presented warning system with a
simple warning system that acts as the baseline. The simple system predicts constant velocity for all agents and uses the resulting risk of the driving situation for the warning signal. We label the novel system as ``human factors'' and the simple system as ``baseline''.

\subsection{Dynamic Lane Change}
The simulation results of the dynamic lane change with no driver error are given in Fig. \ref{fig:warning_no_driver_error}. On the right, two graphs are shown: a graph of the warning signal for both the novel system and the baseline over the simulation time (top) as well as a graph of velocity and acceleration from the motorcycle rider (bottom). On the left, the birds-eye view of the driving situation and the perceived and objective Risk Maps plots of the novel warning system are shown for three timesteps of the simulation. 

\vspace*{-0.00cm}
In the figure, both of the Risk Maps plots are the same. The planned driver behavior from the perceived Risk Maps goes around the risk spot from the objective Risk Maps plot, see the green curve. Therefore, the warning signal stays low. Compared to the baseline model, furthermore, the warning signal stays even lower. The curve ``human factors'' is smaller than ``baseline'' in the warning signal graph. This is because the found driver behavior is avoiding risks and the novel system uses the behavior to estimate the warning signal. 

Without a driver error, no warning is issued and the novel system is more robust than state of the art, such as having less false positive warnings due to a lower warning signal. We can thus set the warning threshold smaller for the novel system, which will become important in the following scenarios with driver errors. The thresholds are set to $w_{\text{thr}}\hspace{-0.02cm}=\hspace{-0.02cm}10^{-4}$ for the novel system and $w^*_{\text{thr}}\hspace{-0.02cm}=10^{-3}$ for the baseline. We selected the thresholds based on a small empirical study of four driving scenarios with no driver errors as input. No false positive warnings were ensured with the chosen thresholds. Additionally, we set them with a safe driver in mind that wants to get warned early.

Fig. \ref{fig:warning_discovery_error} shows an example with a notice error, in which we set ~$\text{NE}_1(t) = 1$ for the simulation run. The motorcycle rider is not aware of the other car in the back. Consequently, the rider is making a lane change but as can be seen in the perceived Risk Maps plot, there is no risk spot predicted and a collision is happening in the last timesteps of the given simulation run. 

\vspace*{-0.1cm}
Since perceived Risk Maps is not detecting risks, the safe driver behavior is to keep constant velocity (compare green curve in perceived Risk Maps). However, the objective Risk Maps plot has a risk spot and the planned behavior goes through the risk spot. The warning signal is accordingly high, which can be observed in the graph of the warning signal. The signal increases over the simulation time. As we can set the threshold lower for the novel warning system from the last experiment without driver errors, we can warn the rider earlier at $3.3$ seconds in the simulation. In comparison, with the baseline model, we warn at $5$ seconds. That means we can warn around 2 seconds earlier in this example with the presented novel warning system. 

Next, we will analyze an example of a forecast error. Fig. \ref{fig:warning_estimation_error} depicts a simulation run of the same scenario with a forecast error of $\text{FE}_1(t)=1$. The rider is estimating the back car's velocity too high. Now, the rider will not change lanes but wait until the car passes. In this case, the perceived Risk Maps plot contains a wide risk spot and the safe driver behavior is to drive on the current lane with constant velocity. However, once the car passes, the rider is making an abrupt lane change behind the car. Here, the behavior change is fast and an earlier warning cannot be achieved. The warning times are similar with $t=12$ seconds in the warning signal graph. 

Finally, Fig. \ref{fig:warning_prediction_error} shows an example with an inference error. The motorcycle rider has an inference error of $\text{IE}_1(t)=1$. Consequently, the rider predicts for the other car a wrong lane change and does not assume that the car stays on its lane. The rider makes a lane change but keeps a too small and slightly unsafe distance to the back car (compare objective Risk Maps visualizations). The found safe driver behavior should be an earlier and stronger acceleration. Due to the difference of perceived and objective Risk Maps, the planned driver behavior goes through a risk spot and the warning signal becomes high. An earlier warning of around 3 seconds can be achieved in this example. 
%To sum up at this point, you saw that the human factors input influence RiskMaps in various ways but how can we now warn and support the driver with this information

\vspace*{-0.05cm}

\subsection{Further Intersection Scenarios}

In order to see if the novel warning system and the concept of driver errors works similarly for further scenarios, such as intersections, we lastly also analyzed the behavior of the system for two intersection cases. 

\begin{figure}[t!]
  \centering
  
  \vspace*{0.2cm}
  
  \resizebox{0.94\linewidth}{!}{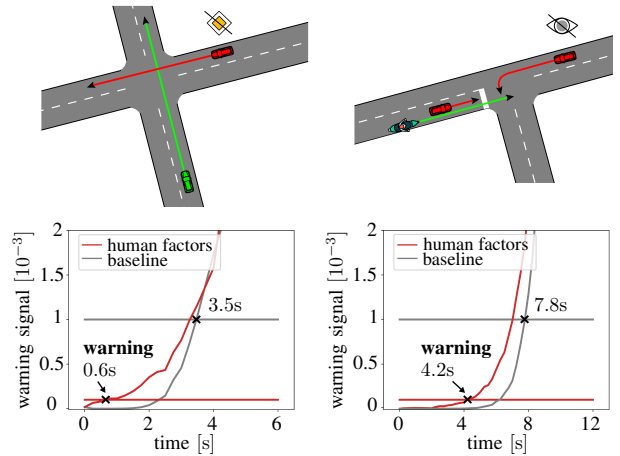}
  
  \vspace*{-0.1cm}
  
  \caption[]{Results of novel warning system on further intersection scenarios. Left: Scenario in which car disobeys priority rules. Right: Scenario in which a car is overlooked by motorcycle rider. In both cases, an earlier warning can be achieved.} 
  
  \vspace*{-0.12cm}
  
  \label{fig:further_examples}
\end{figure}

The first intersection scenario represents a crossing scenario involving two cars. The driver of one car is perceiving the situation wrong because the other car is disobeying the priority rule. The second intersection scenario is a complex scenario, in which a motorcycle rider overlooks a car because the view is obstructed by a further car from the point of the rider.
A snapshot of the driving situations and the course of the warning signal are shown in Fig. \ref{fig:further_examples}.

In the case of the disobeying car behavior, the driver has a forecast error of $\text{FE}(t) = 1$. The driver assumes the other car to drive with low velocity, although the car drives with high velocity. As can be seen in the plot of the warning signal, by detecting the forecast error, an earlier warning of $2.9$ seconds can be achieved. Similarly, for the case of the overlooked car, a detection of a notice error of $\text{NE}(t) = 1$ allows to warn the motorcycle rider $3.6$ seconds earlier of the upcoming dangerous car. This shows that the presented driver errors and novel warning system can be flexibly applied to many different driving scenarios.

\subsection{Discussion}
In this last section, we will shortly discuss the limitations of the experiments and explain possible improvements for future work. As described,  the warning thresholds for both the novel system and the baseline system were chosen based on a small empirical study with no driver error inputs. The study included two following scenarios and two intersection scenarios with maximum three vehicles. In order to refine these thresholds, a larger set of driving scenarios is needed. In this large-scale study, the systems should output only few or no false positives with the chosen thresholds. In addition, the human factors input for the system were ideal in the experiments. Human factors input with wrong estimations can lead to a reduced warning time. An accurate and reliable human state estimation module is a requirement for the novel system. Overall, with these limitations in mind, the results of the experiments from this paper  show that human factors allow to improve the driver support effectively with Risk Maps.

\section{Conclusion and Outlook}
\label{sec:conclusion}

In summary, in this paper, we proposed a novel warning system concept that considers human factors in the form of three driver errors, which are a) notice error, b) forecast error and c) inference error. The system accordingly plans a driver behavior based on Risk Maps that includes the driver errors and checks if the planned behavior is objectively safe. This warning concept highlights the opportunity to bridge the important gap between human factors research and driver support systems. In comparison, previous work focused only either on the personalization of the driver support based on driving style or on perceived risks with human-centered risk sources, such as looking at the smartphone. 

In different simulations of lane change and of intersection scenarios, the novel warning system was finally shown to be less prone to false positive warnings and was able to warn the driver in some cases several seconds earlier than state of the art that does not include human factors. Considering human factors for driver warning with Risk Maps therefore allowed for more robust and foresighted driver warning. 

In future work, we target to further improve the support with human factors. For example, in one of the experiments, the warning could not be improved because of an abrupt change of driver behavior. We target to find out how human factors can improve driver supports in these more short-term 
cases. One possibility is to improve the behavior prediction for the vehicles, as it was done, e.g., in \cite{jain2015}. In this way, the rapid change might be detected earlier, and a warning can be given before the action is done.

Furthermore, we target to apply the novel warning system \makebox[1.01\linewidth][s]{on a real test vehicle with all needed components integrated.} 

\newpage 
\noindent In the presented simulations, the state estimation module was not implemented and the human factors input was given. By combining a state estimation module with the warning system, the impact of noise in the input can be investigated. Here, the general effectiveness of a chosen Human-Machine Interface (HMI) that communicates the warning needs to also be investigated.

\addtolength{\textheight}{-12cm}   % This command serves to balance the column lengths
                
%\vspace*{0.01cm}

\bibliographystyle{IEEEtran}
%{\footnotesize \bibliography{bib.bib}}
\bibliography{bib}

\begin{thebibliography}{10}
\providecommand{\url}[1]{#1}
\csname url@samestyle\endcsname
\providecommand{\newblock}{\relax}
\providecommand{\bibinfo}[2]{#2}
\providecommand{\BIBentrySTDinterwordspacing}{\spaceskip=0pt\relax}
\providecommand{\BIBentryALTinterwordstretchfactor}{4}
\providecommand{\BIBentryALTinterwordspacing}{\spaceskip=\fontdimen2\font plus
\BIBentryALTinterwordstretchfactor\fontdimen3\font minus
  \fontdimen4\font\relax}
\providecommand{\BIBforeignlanguage}[2]{{%
\expandafter\ifx\csname l@#1\endcsname\relax
\typeout{** WARNING: IEEEtran.bst: No hyphenation pattern has been}%
\typeout{** loaded for the language `#1'. Using the pattern for}%
\typeout{** the default language instead.}%
\else
\language=\csname l@#1\endcsname
\fi
#2}}
\providecommand{\BIBdecl}{\relax}
\BIBdecl

\bibitem{bengler2014}
K.~Bengler, K.~Dietmayer, B.~Farber, M.~Maurer, C.~Stiller, and H.~Winner,
  ``{Three Decades of Driver Assistance Systems: Review and Future
  Perspectives},'' \emph{IEEE Intelligent Transportation Systems Magazine},
  vol.~6, no.~4, pp. 6--22, 2014.

\bibitem{bianchi2013}
G.~Bianchi~Piccinini, G.~Prati, L.~Pietrantoni, C.~Manzini, C.~Rodrigues, and
  M.~Leitao, ``{Drivers' hand positions on the steering wheel while using
  Adaptive Cruise Control (ACC) and driving without the system},'' in \emph{The
  Human Factors and Ergonomics Society, Europe Chapter Conference}, 2013.

\bibitem{zhu2021}
H.~Zhu, T.~Misu, S.~Martin, X.~Wu, and K.~Akash, ``{Improving Driver Situation
  Awareness Prediction using Human Visual Sensory and Memory Mechanism},'' in
  \emph{IEEE/RSJ International Conference on Intelligent Robots and Systems},
  2021, pp. 6210--6216.

\bibitem{hasenjaeger2017}
M.~Hasenj\"{a}ger and H.~Wersing, ``Personalization in advanced driver
  assistance systems and autonomous vehicles: A review,'' in \emph{IEEE
  International Conference on Intelligent Transportation Systems}, 2017, pp.
  1--7.

\bibitem{siebinga2022}
O.~Siebinga, A.~Zgonnikov, and D.~Abbink, ``{A Human Factors Approach to
  Validating Driver Models for Interaction-Aware Automated Vehicles},''
  \emph{ACM Transactions on Human-Robot Interaction}, vol.~11, no.~4, 2022.

\bibitem{puphal2019}
T.~Puphal, M.~Probst, and J.~Eggert, ``{Probabilistic Uncertainty-Aware Risk
  Spot Detector for Naturalistic Driving},'' \emph{IEEE Transactions on
  Intelligent Vehicles}, vol.~4, no.~3, pp. 406--415, 2019.

\bibitem{kolekar2021}
S.~Kolekar, B.~Petermeijer, E.~Boer, J.~{de Winter}, and D.~Abbink, ``A risk
  field-based metric correlates with driver's perceived risk in manual and
  automated driving: A test-track study,'' \emph{Transportation Research Part
  C: Emerging Technologies}, vol. 133, 2021.

\bibitem{charlton2014}
S.~Charlton, N.~Starkey, J.~Perrone, and R.~Isler, ``{What's the risk? A
  comparison of actual and perceived driving risk},'' \emph{Transportation
  Research Part F: Traffic Psychology and Behaviour}, vol.~25, pp. 50--64,
  2014.

\bibitem{fridman2018}
L.~Fridman, ``Human-centered autonomous vehicle systems: Principles of
  effective shared autonomy,'' \emph{arXiv}, 2018.

\bibitem{wang2020}
C.~Wang, M.~Kr\"{u}ger, and C.~B. Wiebel-Herboth, ``{"Watch out!":
  Prediction-Level Intervention for Automated Driving},'' in
  \emph{International Conference on Automotive User Interfaces and Interactive
  Vehicular Applications}, 2020, pp. 169--180.

\bibitem{mccall2007}
J.~C. McCall and M.~M. Trivedi, ``{Driver Behavior and Situation Aware Brake
  Assistance for Intelligent Vehicles},'' \emph{Proceedings of the IEEE},
  vol.~95, no.~2, pp. 374--387, 2007.

\bibitem{puphal2022}
T.~Puphal, B.~Flade, M.~Probst, V.~Willert, A.~J\"urgen, and J.~Eggert,
  ``{Online and Predictive Warning System for Forced Lane Changes Using Risk
  Maps},'' \emph{IEEE Transactions on Intelligent Vehicles}, vol.~7, no.~3, pp.
  616--626, 2022.

\bibitem{jain2015}
A.~Jain, H.~Koppula, B.~Raghavan, S.~Soh, and A.~Saxena, ``{Car that Knows
  Before You Do: Anticipating Maneuvers via Learning Temporal Driving
  Models},'' in \emph{IEEE International Conference on Computer Vision}, 2015,
  pp. 3182--3190.

\end{thebibliography}

\end{document}